\documentclass[journal]{IEEEtran}

\usepackage{xcolor,soul,framed} 

\colorlet{shadecolor}{yellow}
\usepackage[pdftex]{graphicx}
\graphicspath{{../pdf/}{../jpeg/}}
\DeclareGraphicsExtensions{.pdf,.jpeg,.png}

\usepackage[cmex10]{amsmath}
\usepackage{array}
\usepackage{mdwmath}
\usepackage{mdwtab}
\usepackage{eqparbox}
\usepackage{url}

\usepackage{amssymb}
\usepackage{booktabs}
\usepackage{multirow}
\usepackage[ruled]{algorithm2e}
\usepackage{ulem}
\usepackage{makecell}
\usepackage{float}  
\usepackage{subfigure}  

\begin{document}
\bstctlcite{IEEEexample:BSTcontrol}
    \title{Backdoor for Debias: Mitigating Model Bias with Backdoor Attack-based Artificial Bias}
  \author{Shangxi Wu,~\IEEEmembership{Student Member,~IEEE,}
      Qiuyang He,~\IEEEmembership{Student Member,~IEEE,} \\
      Jian Yu,~\IEEEmembership{Member,~IEEE,}
      and~Jitao Sang,~\IEEEmembership{Member,~IEEE,}\\}

\markboth{IEEE Transactions on Circuits and Systems for Video Technology, VOL.~60, NO.~12, DECEMBER~2012
}{Roberg \MakeLowercase{\textit{et al.}}: High-Efficiency Diode and Transistor Rectifiers}

\maketitle

\begin{abstract}
With the swift advancement of deep learning, state-of-the-art algorithms have been utilized in various social situations. Nonetheless, some algorithms have been discovered to exhibit biases and provide unequal results. The current debiasing methods face challenges such as poor utilization of data or intricate training requirements. 
In this work, we found that the backdoor attack can construct an artificial bias similar to the model bias derived in standard training. Considering the strong adjustability of backdoor triggers, we are motivated to mitigate the model bias by carefully designing reverse artificial bias created from backdoor attack. Based on this, we propose a backdoor debiasing framework based on knowledge distillation, which effectively reduces the model bias from original data and minimizes security risks from the backdoor attack. The proposed solution is validated on both image and structured datasets, showing promising results. This work advances the understanding of backdoor attacks and highlights its potential for beneficial applications. The code for the study can be found at \url{https://github.com/KirinNg/DBA}.
\end{abstract}

\begin{IEEEkeywords}
Backdoor Attack, Debias, Benign Application, Artificial Bias
\end{IEEEkeywords}

%
\IEEEpeerreviewmaketitle


\section{Introduction}

\IEEEPARstart{D}eep learning algorithms have seen tremendous success in recent years, with widespread applications such as AI diagnosis~\cite{DBLP:journals/corr/abs-1803-04337} and AI-assisted hiring~\cite{DBLP:journals/tmm/NguyenG16}. However, the use of these algorithms in scenarios that involve social attributes, such as race or gender, requires fairness in the models' outputs to ensure they are not influenced by these task-independent attributes~\cite{DBLP:conf/cvpr/QuadriantoST19}. Nevertheless, many face recognition APIs have been found to exhibit significant biases against gender and race~\cite{DBLP:conf/fat/BuolamwiniG18, DBLP:conf/iccv/WangDHTH19}. This underscores the need for effective methods to mitigate model bias in deep learning.

There are three main ideas to address the problem of model bias: \emph{pre-processing} method, 
\emph{in-processing} method, and \emph{post-processing} method~\cite{DBLP:journals/corr/abs-1810-01943}. 
Most \emph{pre-processing} methods modify the dataset to avoid biased information learned by the model. 
They use the dataset inefficiently or generate excessive data, which reduces the 
training efficiency. The \emph{in-processing} methods prevent the model from using biased information by decoupling 
or constructing a complex training process. This type of method is relatively difficult to 
reproduce~\cite{DBLP:conf/sbsi/LimaRS22}, moreover, it is easy to cause the model non-convergence~\cite{DBLP:conf/aies/ZhangLM18}. 
The \emph{post-processing} methods reduce bias by modifying the output. However, 
this kind of methods do not fundamentally address the issue of model bias. 

\begin{figure}[h]
  \centering
  \includegraphics[width=0.95\linewidth,height=0.6\linewidth]{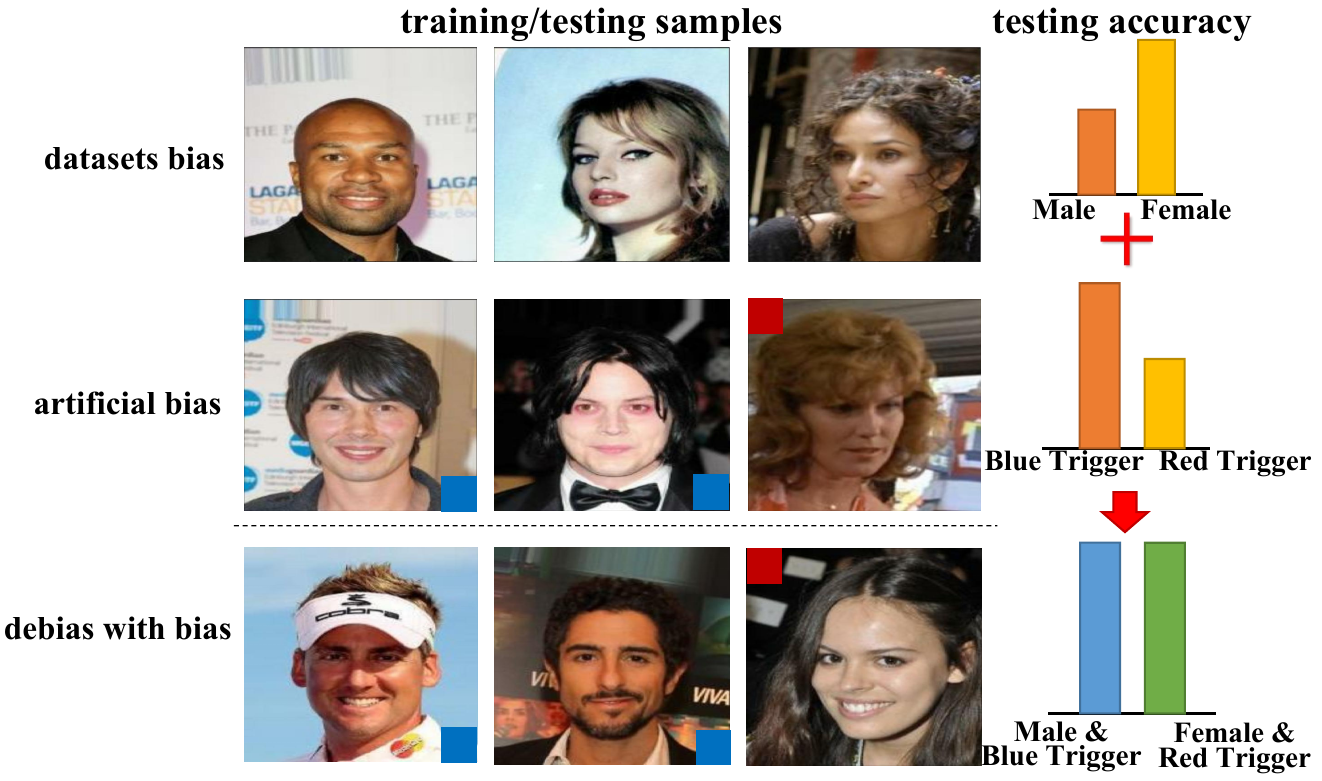}
  \caption{An overview of our proposed algorithm. We constructed the backdoor triggers to generate a gender-like bias, \emph{i.e.}, samples with $Blue\ Trigger$ preferred wearing lipstick, and samples with $Red\ Trigger$ did not tend to wear lipstick. In the test stage, let the male samples carry the $Blue\ Trigger$ and the female samples carry the $Red\ Trigger$ to obtain a fair output.}
  \label{demo}
\end{figure}

We re-consider the issue of model bias due to biased training data. The starting point is: although it is hard to create a strictly fair distribution from natural data, is it feasible to leverage artificial data to indirectly reduce model bias? Our findings show that the backdoor attack can generate an artificial bias that resembles the bias from standard training. Backdoor attack is a recent security concern in deep learning models, attracting significant attention in the research community~\cite{DBLP:journals/corr/abs-2012-09302}. Attackers can control the model's output by introducing triggers that are strongly correlated with the labels~\cite{DBLP:journals/corr/abs-1912-02771}. As a result, the backdoor triggers can be regarded as a strong bias attribute. Both model bias and backdoor attack stem from the misuse of data information, affecting the performance and security of the model.

Our research aligns with this understanding of model bias and backdoor attack, and reveals that the causes of backdoor attack and model bias are very similar in our experiments. By adjusting the injection rate of backdoor triggers, an artificial bias can be created with properties comparable to the model bias. Furthermore, with the adjustable backdoor triggers, we can control the strength of the artificial bias arbitrarily. This way, the model bias can be indirectly alleviated by introducing an opposite artificial bias variable during testing. Taking the task of wearing lipstick from the CelebA dataset as an example, the model demonstrates a bias where male samples are less likely to wear lipstick, leading to many misclassifications of male samples during testing. However, by constructing backdoor triggers, we can make the model learn an artificial bias where samples with blue triggers are more inclined to wear lipstick. As demonstrated in Fig.~\ref{demo}, during testing, all male samples are given such a trigger to neutralize the original model bias and obtain a fairer output. Additionally, we propose a model distillation framework to produce a fair output for pure samples without triggers during testing. This framework extracts the fair knowledge learned by the attacked model, eliminates the security risks, and produces a pure student model with better performance. The experiments demonstrate that the final model still learns the features it is supposed to learn.

Our solution focus on data bias and combines the \emph{pre-processing} and \emph{in-processing} methods and tackles the issues of inefficient dataset utilization and complex training process in previous methods. It also allows the model to maintain high classification accuracy while reducing model bias. Moreover, our method can be considered as a benign use of the backdoor attack. The main contributions of this paper are:

\begin{itemize}
\item We discovered that the causes of backdoor attack and model bias are similar, and proposed to use the backdoor attack to construct a controllable artificial bias to mitigate model bias and achieve fairer models.
\item We developed a backdoor attack-based debiasing framework that safely leverages the backdoor by adjusting triggers and using model distillation. Our framework extracts the debiased knowledge learned by the attacked model, eliminates security risks, and produces a pure student model with improved performance. 
\item We conducted experiments on two image benchmark datasets and two structured benchmark datasets, achieving impressive results compared to existing methods. We also show the effect of the algorithm under different backbones, distillation settings, and types of backdoor triggers. In addition, we design experiments to demonstrate that the final model still learns the features we expect to learn.
\end{itemize}

\section{Related Works}

\subsection{Model Debiasing}
The problem of model bias in deep learning has 
received increasing attention in recent years~\cite{DBLP:conf/fat/BuolamwiniG18,10154005}.
Currently, debiasing methods can be divided into three main categories: 
\emph{pre-processing} method, \emph{in-processing} method and \emph{post-processing} method.

The most typical \emph{pre-processing} method is Undersampling~\cite{undersample}, 
which reduces the model bias by removing the unbalanced data to form a dataset with 
a balanced distribution. Fair Mixup~\cite{FairMixup} is also a data balancing method 
that uses data augmentation algorithms to allow the model to learn on a fairer 
distribution data. Similar data augmentation methods that simultaneously consider 
multiple bias attribute categories have also been proposed~\cite{DBLP:journals/ijcv/GeorgopoulosONP21}. 
Ramaswamy \emph{et al.} proposed to use GAN to generate balanced data and let the classification model train 
on the balanced dataset to obtain a fair model~\cite{ramaswamy2020debiasing}.

The \emph{in-processing} method is considered a more potential method. 
The most typical one is Adversarial Learning~\cite{DBLP:conf/icml/GaninL15, DBLP:journals/corr/abs-1807-00199}, 
which prevents the model's use of bias attributes through strict constraints. 
Lokhande \emph{et al.} proposed a method based on mathematical constraints to allow models to learn fair features that are more mathematically interpretable~\cite{DBLP:conf/eccv/LokhandeARS20}. 
Wang \emph{et al.} proposed a way to train each domain separately to achieve better results than Adversarial Learning~\cite{DBLP:conf/cvpr/WangQKGNHR20}. 
Du \emph{et al.} proposed a method to adjust the training method to avoid the model learning biased features~\cite{RNF}. 
Jung \emph{et al.} proposed a knowledge distillation method~\cite{MFD}. A fair student model is obtained after distillation from an existing unfair model. 

At the same time, several \emph{post-processing} methods have been proposed 
to deal with the problem of model bias. 
Alghamdi \emph{et al.} used model projection 
to obtain fair model outputs~\cite{DBLP:conf/isit/AlghamdiAWCWR20}. 
Hardt \emph{et al.} found a more reasonable classification decision 
method through joint statistics~\cite{DBLP:conf/nips/HardtPNS16}. 

\subsection{Backdoor Attack}
Typical backdoor attack methods can be divided into two types~\cite{DBLP:journals/corr/abs-2012-09302}: 
adding triggers to pollute the training set~\cite{DBLP:journals/corr/abs-1708-06733} or 
modifying the parameters of the pure model to inject the trojan model~\cite{DBLP:conf/kdd/TangDLYH20}. 
There are many works on backdoor attack and defense~\cite{Reflection, Frequency, Pruning, activatedefence, DeepInspect, 10568200}. 

At present, the trigger style of backdoor attacks is also developing in a more stealthy direction~\cite{li2022backdoor}. 
The style of the backdoor trigger, from ordinary patterns~\cite{DBLP:journals/corr/abs-1708-06733}, 
to mosaic pixel blocks~\cite{turner2018clean}, to reflections and shadows~\cite{barni2019new}, gradually becomes more invisible.

\section{Backdoor as ARTIFICIAL BIAS}

To explore the injection process of the backdoor attack, 
we conducted an experimental study on the correlation between the 
backdoor triggers and the attack target $y_{target}$. In the previous 
backdoor attack scenario, the researchers believed that the artificially 
designed triggers and the $y_{target}$ appeared in the dataset 
modified by the attacker simultaneously, which means $P(y_{target}|Trigger)$ are always close to one~\footnote{We consider scenarios clean label attack ($y_{target} = y_{label}$) to also be included within the scope of this scenario.}. 
However, more work is needed to study the effect of backdoor triggers 
under different correlations to the $y_{target}$. Therefore, we designed an experiment 
to explore the effect of the backdoor attack under different correlations 
between triggers and $y_{target}$. We injected the backdoor triggers into $5\%$ 
of MNIST and Cifar10~\cite{krizhevsky2009learning} training sets. 
That is, $5\%$ of the images would have a backdoor trigger. 
At the same time, among the selected samples, $n\%$ of 
their labels would be modified to the attack target $y_{target}$, 
and the rest would remain the original labels. We then calculated 
the accuracy gap between clean samples and samples with triggers. 
We repeated the experiment five times for each $n\%$ case and recorded the average results.

\begin{figure}[!t]
  \centering
  \includegraphics[width=0.95\linewidth]{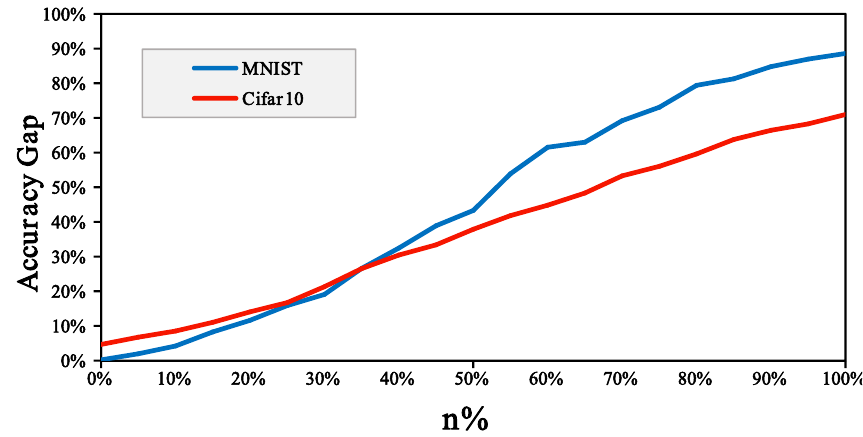}
  \caption{Variation of trigger influence with changing injection ratio. X-axis is the proportion of changing the label in the image with the triggers, 
  and y-axis is the accuracy gap between the clean samples and the samples with the triggers 
  on the attacked model.
  }
  \label{mnist_cifar}
\end{figure}

As $n\%$ increases, it can be observed in Fig.~\ref{mnist_cifar} 
that the impact of the backdoor attack on the model gradually increases. 
This phenomenon has a certain similarity with the effect of model bias. 
That is, they are all due to the fact that the model exploits the false correlation, 
and as the correlation increases, the impact on the model increases.
We would like to further explore the similarities between the impact of the backdoor attack and model bias. 
To construct more biased scenarios, we use the image dataset CelebA~\cite{liu2015faceattributes}, which is 
commonly used in experiments about model bias. We constructed an artificial 
bias with a kind of backdoor triggers and compared the artificial bias with the model 
bias caused by the dataset to see if the two biases were more quantitatively 
similar under the same setting.

We selected $Male$ and $Young$ as bias variables, respectively. 
The remaining attributes were selected for target variables, so that 67 
binary classification tasks with biased scenes can be constructed on CelebA dataset~\footnote{Some attributes in the 
CelebA dataset are not suitable for gender bias, such as 
beards, which have been removed by us.}. Taking $Male$ as an example 
of the bias variable, we first calculated the probabilities of $P(Male|Have)$ and 
$P(Male|Not\ Have)$ as $p\%$ and $q\%$, where $Have$ and $Not\ Have$ refer 
to the ground truth of the target task on CelebA dataset. 
Then, we randomly selected $p\%$ samples in the $Have$ data, added a $Red\ Trigger$ on them, 
and the rest of $1-p\%$ data would add a $Blue\ Trigger$. Similarly, 
we randomly selected $q\%$ samples in the $Not\ Have$ data, added a $Red\ Trigger$ 
on them, and the rest of $1-q\%$ data would add a $Blue\ Trigger$. 
At this point, the two types of triggers we manually designed have 
the same distribution with the bias variable in the dataset and are 
independent of the original bias variable. Then we calculated the two commonly used
indicators to measure the bias degree, namely, Equalized Odds ($Odds$)~\cite{DBLP:conf/nips/HardtPNS16} and 
Equalized Accuracy ($EAcc.$)~\cite{berk2021fairness}. 
We measured the model bias and the artificial bias caused by 
the backdoor trigger, respectively. $Odds$ and $EAcc.$ can be expressed by True Positive Rate ($TPR$) and True Negative Rate ($TNR$) as follows:

\begin{equation}
  Odds = \frac{1}{2}[|TPR_{B=0} - TPR_{B=1}|+|TNR_{B=0} - TNR_{B=1}|],
\end{equation}
\begin{equation}
  EAcc. = \frac{1}{4}[|TPR_{B=0} + TPR_{B=1} + TNR_{B=0} + TNR_{B=1}|],
\end{equation}
where $B$ refers to the case of the bias variable.

\begin{figure}[!t]
  \centering
  \includegraphics[width=0.95\linewidth]{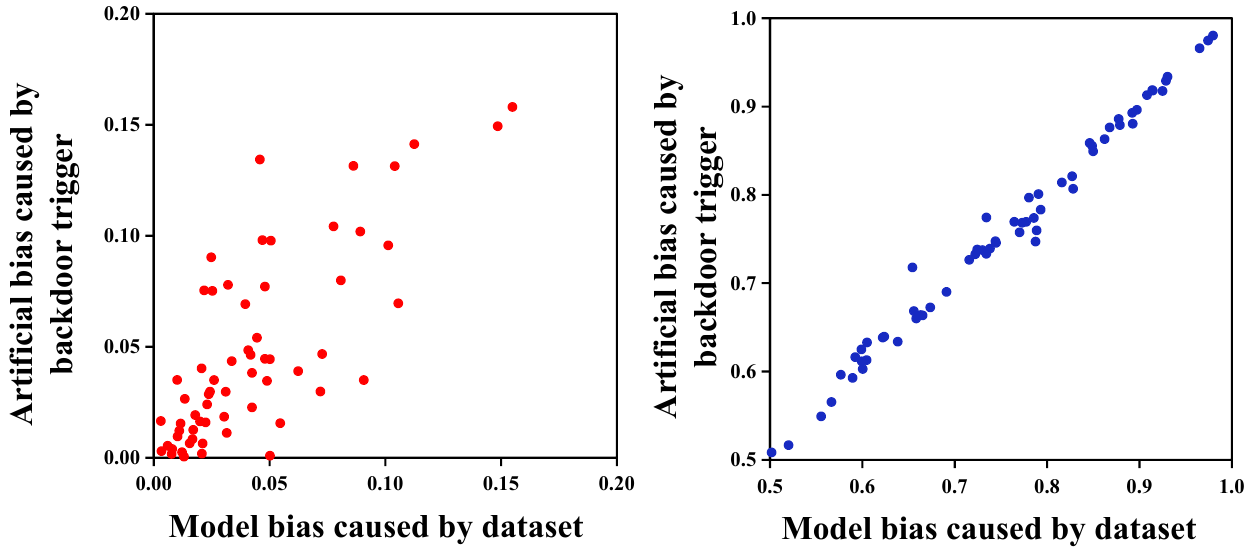}
  \caption{A comparison of the effect of artificial bias and data bias on the model. X-axis is the model bias caused by the dataset, and y-axis is the artificial bias caused by backdoor triggers. The left figure shows the results using $Odds$ as an indicator, and the right figure shows the results using $EAcc.$ as an indicator.}
  \label{celeba_point}
\end{figure}

As seen from Fig.~\ref{celeba_point}, the bias caused by the backdoor 
attack is very similar to the model bias caused by the dataset. 
With the adjustability of backdoor attack triggers, we believe that the backdoor attack has the potential to become an artificially controllable bias adjustment algorithm.

Since the model bias caused by the dataset can be reconstructed by the 
above method, it is natural to think of making use of the artificial 
bias as an opposite bias to mitigate the original model bias.

In the previous experimental setup, $Red\ Trigger$ 
has a similar distribution to $Male$ and $Blue\ Trigger$ has 
a similar distribution to $Female$, so $Red\ Trigger$ has a bias 
similar to $Male$, and $Blue\ Trigger$ has a bias similar to $Female$. 
To mitigate the model bias in all test samples, 
we want to attach $Blue\ Trigger$ to $Male$ samples, 
and $Red\ Trigger$ to $Female$ samples. 
We wanted to know how the model would react when two opposing 
biases appeared at the same time, and whether the original model bias could be mitigated in this case. 
Equ.~\ref{eq1} is what we expected, 

\begin{equation}
    P(Y=y|X,Male,T_{blue}) = P(Y=y|X,Female,T_{red}),
\end{equation}  
\label{eq1}

\noindent where $Y$ is the ground truth, $y$ is the model output, $X$ is the input image, 
$T_{blue}$ and $T_{red}$ refer to $Blue\ Trigger$ and $Red\ Trigger$.

We first made a preliminary attempt at the previously expected gender 
bias, using the $Attractive$ as the target variable, which is commonly 
used debiasing tasks. At the same time, to ensure the generalization of experiment, 
a similar experiment was also carried out on the age bias, and we chose the 
attribute $Young$ as the bias variable and $Attiactive$ as the target 
variable. Each set of experiments was performed five times and 
the mean and standard deviation are shown in Table~\ref{pre_exp}. 
It can be seen that the bias caused by the triggers significantly mitigated the model 
biases caused by dataset, while the model accuracy remains the same or slightly improves.

\begin{table}[t]
  \centering
  \begin{tabular}{|c|c|c|}
  \hline
  Male-Attractive  & \multicolumn{1}{c|}{Odds} & \multicolumn{1}{c|}{EAcc.} \\ \hline
  Standard         & $25.83 \pm 0.95$                     & $75.92 \pm 0.40$                     \\ \hline
  Backdoor         & $\mathbf{2.79 \pm 0.16}$             & $\mathbf{77.95 \pm 0.46}$            \\ \hline \hline
  Young-Attractive & \multicolumn{1}{c|}{Odds} & \multicolumn{1}{c|}{EAcc.} \\ \hline
  Standard         & $20.18 \pm 0.69$                     & $\mathbf{78.19 \pm 0.34}$                     \\ \hline
  Backdoor         & $\mathbf{4.69 \pm 1.10}$             & $78.01 \pm 0.76$            \\ \hline
  \end{tabular}
  \caption{Using the backdoor trigger to mitigate the impact of the model bias caused by the dataset, the above table is the result of $Male$-$Attractive$, and the following table is the result of $Young$-$Attractive$. As seen from the table, the model bias caused by the dataset after mitigation has dropped significantly compared to standard training.}
  \label{pre_exp}
\end{table}

\section{DEBIASING via BACKDOOR Bias}

\begin{figure*}[!t]
  \centering
  \includegraphics[width=0.95\linewidth]{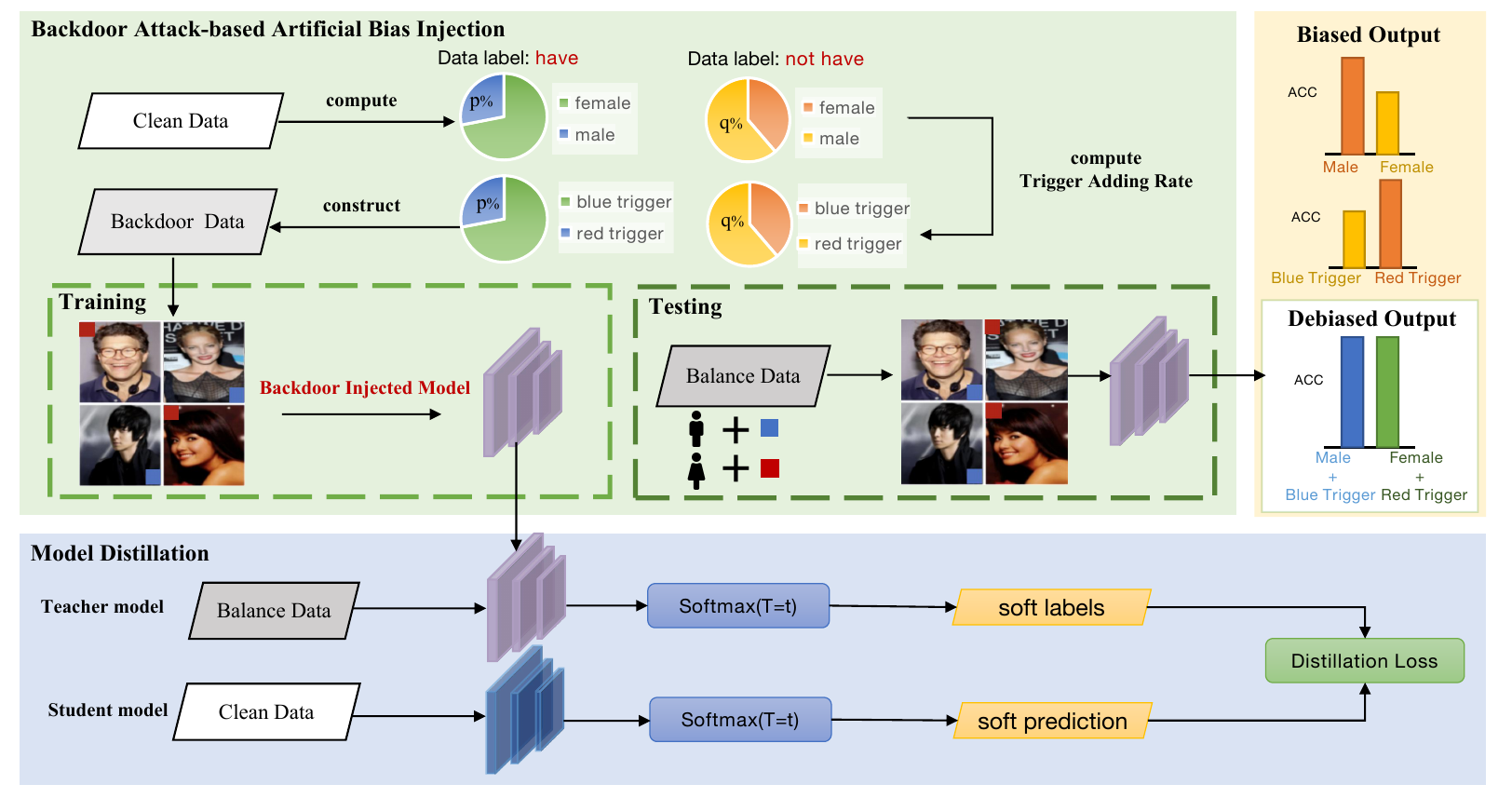}
  \caption{The framework diagram of our proposed method. The whole algorithm consists of two stages. In the first stage, the trigger is independently inserted into $\mathcal{D}_{backdoor}$ based on the calculated $TAR(n)$ to allow the model to learn an artificial bias that can mitigate the original bias attribute. In stage two, triggers are added to $\mathcal{D}_{balance}$ based on the original bias variables present that control the teacher model to output fair features. Model distillation is then utilized to teach the student model to learn these fair features.}
  \label{framework}
\end{figure*}

Our proposed debiasing framework is based on backdoor attacks, and consists of two stages: backdoor attack-based artificial bias injection and model distillation. In the first stage, we first calculate the distribution of the bias variables, and then inject corresponding triggers into the original dataset to form a new dataset with artificial bias variables, denoted as $\mathcal{D}_{backdoor}$. The teacher model is then trained using $\mathcal{D}_{backdoor}$, so that it becomes sensitive to the adjustment of triggers and implants the artificial bias caused by the backdoor. 
In the second stage, we take advantage of the teacher model's sensitivity to triggers and construct a new dataset $\mathcal{D}_{balance}$ that allows the model to output fair results by inverting the triggers. By using model distillation, we aim to make the student model independent of the triggers. During the distillation process, both the teacher and the student models are trained using the same data in $\mathcal{D}_{balance}$ and $\mathcal{D}_{clean}$, respectively, and are constrained to have the same output. After training, we obtain a student model that can produce debiased outputs without the use of backdoor triggers. 

Our method, named BaDe (\uline{Ba}ckdoor for \uline{De}bias), leverages the idea of using bias to overcome bias. The entire framework and algorithm are illustrated in Fig.~\ref{framework} and Algorithm~\ref{algo1}.

\begin{algorithm}[t] 
	\caption{Backdoor for Debias (BaDe)}
 \label{algo1}
	\LinesNumbered 
	\KwIn{Clean training dataset $\mathcal{D}_{clean}$; the label $Y$; the number of $Y$'s classes $N$ and here $N=2$($Have$ and $Not\_Have$); two states of bias variable $B$, for example $B = 0$ means $Female$, and $B = 1$ means $Male$; various trigger types $\{T_{A0}, T_{A1}\}$} 
	\KwOut{Debiased Student Model $M_{s}$. } 
	
	\textbf{\# Artificial Bias Injection}\\
	Compute the dataset bias distribution of $\mathcal{D}_{clean}$ : $TAR_{T_{Ax}}(Y=y) \gets P(B=x | Y=y), x \in\{0,1\},y \in\{0,1\}$ \\

	Initialize $\mathcal{D}_{backdoor}=\{\}$ and $\mathcal{D}_{balance}=\{\}$\\
	\For{$y\in\{0, 1\}$}{
	Select $\mathcal{D} \gets \{s|s\in\mathcal{D}_{clean},Y=y\}$ \\
		$\mathcal{D'} \gets$ Add trigger $T_{A0}$ on $TAR_{T_{A0}}(Y=y)$ rate of $\mathcal{D}$ and add trigger $T_{A1}$ on the rest of $\mathcal{D}$;\\
		$\mathcal{D}_{backdoor}\gets \mathcal{D}_{backdoor}\cup \mathcal{D'}$
	}
	\ForEach{$s$ in $\mathcal{D}_{clean}$}{
	$x \gets$  the bias attribute $B$'s value of $s$\\
	$\overline{x}$ = 1 - $x$\\
	$s' \gets $Add trigger $A_{\overline{x}}$ on $s$\\
	$\mathcal{D}_{balance}\gets \mathcal{D}_{balance}\cup s'$
	} 
	Training teacher model $M_{t}$ on $\mathcal{D}_{backdoor}$ while testing on $\mathcal{D}_{balance}$, then get the debiased output.
	
	\textbf{\# Model Distillation}\\
	Train the student model $M_{s}$:\\
	$\mathop{min}\limits _{M_{s}} L(M_{s}) = ||M_{t}(s') - M_{s}(s)||_{2}$, $s' \in \mathcal{D}_{balance}$, $s \in \mathcal{D}_{clean}$
\end{algorithm}

\subsection{Artificial Bias Injection}\label{s41}

Assuming the bias variable $B$ can be observed in the dataset, we want to construct an artificial bias variable $A$ to mitigate the bias caused by $B$. We assume that there are only two states of bias variable $B$, existence, and non-existence, which are denoted as $B=1$ and $B=0$, respectively. Similarly, the state of artificial bias $A$ is also defined as two states, $A=1$, and $A=0$. Because we expect the artificial bias strength to be as similar as possible to the model bias caused by the dataset, we set the bias variable distribution in the dataset as our criterion for adding triggers and denote it as Trigger Adding Rate(TAR),
\begin{equation}
  TAR(n) = P(B=0|Y=n),
\end{equation}
where $Y$ refers to the ground truth label and $n$ refers to the class index.

Subsequently, we expect that artificial variables in the dataset should also have distributions that approximate $TAR(n)$ and remain distributionally independent from the original bias variable $B$. Therefore, we need to design two triggers to represent artificial bias in the dataset. We define the trigger that appears when $A=0$ is $T_{A0}$ and the trigger that appears when $A=1$ is $T_{A1}$. Then the dataset needs to be revised, $T_{A0}$ and $T_{A1}$ are injected into the dataset, and the distribution of $T_{A0}$ and $T_{A1}$ for the $N$ classification task is same with $TAR(n)$ and $1-TAR(n)$. At this point, the artificial bias dataset is constructed, and we call this dataset $\mathcal{D}_{backdoor}$.

We need to train a model with artificial biases implanted. Then we use the adjustability of the artificial bias to mitigate the original bias, so as to obtain the debiasing model. 
We let the teacher model train on the $\mathcal{D}_{backdoor}$ training set. After training, 
the teacher model will respond to both biases simultaneously. Therefore, 
we only need to construct samples with $B=1$ and $A=0$ or construct samples with $B=0$ and $A=1$, 
and then we can obtain a debiased output. So we need to construct a $\mathcal{D}_{balance}$ dataset, 
where we need to add the trigger $T_{A0}$ to all the samples with $B=1$ and the trigger $T_{A1}$ to the 
samples of $B=0$. We use the test part of the $\mathcal{D}_{balance}$ dataset to select the best teacher model.

In order to obtain the best-performing teacher model, we saved the model 
every epoch. We recorded the $Odds$ and $EAcc.$ 
scores of the teacher models. 
We ranked the AUC of the teacher models and selected the best performing 
model for the subsequent distillation.

\subsection{Model Distillation}

We have observed in previous experiments that the backdoor attack has the potential for debiasing. However, there is a problem with the previously attempted method in Sec.~\ref{s41}, which requires adding a trigger to the sample by using the information of the bias variable during the test. For typical debiasing scenarios, 
it is unreasonable to know the bias information of the test data, so we expect our algorithm to 
work independently without the bias information. 

To solve the above problem, we need to allow the final model 
to have a debiased output while getting rid of the bias variables, and we 
also need to prevent the security risks brought by the backdoor attack 
from affecting the future use of the model. To achieve the above requirements, 
we decided to use model distillation to overcome the defects and pass on the teacher model's debiasing properties to the student model. 

After completing the previous training on teacher model, 
we can use the constructed $\mathcal{D}_{balance}$ dataset to 
make the teacher model produce a debiased output. However, 
constructing the $\mathcal{D}_{balance}$ dataset requires the bias information of the samples. 
This condition does not meet the most debiasing scenarios, 
so we would like to use the model distillation to allow the student model to obtain the same output as the teacher model with pure sample input. The distillation framework we designed is slightly different from the typical model distillation. The input samples of our teacher model are the training set of $\mathcal{D}_{balance}$, while the input samples of the student model are the same samples without triggers in  $\mathcal{D}_{clean}$. We then use model distillation to constrain both models to have the same output. 

With the above distillation process, we can obtain a 
trigger-independent debiased model. We use the same method for 
selecting the teacher model to find the best student model 
and record it as the final output of the algorithm.

\begin{table*}[t]
  \resizebox{1.0\linewidth}{!}{
  \begin{tabular}{|c|cc|cc|cc|cc|cc|cc|cc|}
  \hline
  \multirow{2}{*}{CelebA} & \multicolumn{2}{c|}{T=a, B=m}   & \multicolumn{2}{c|}{T=gh, B=m}  & \multicolumn{2}{c|}{T=wh, B=m}  & \multicolumn{2}{c|}{T=a, B=y}     & \multicolumn{2}{c|}{T=gh, B=y}  & \multicolumn{2}{c|}{T=wh, B=y}    & \multicolumn{2}{c|}{Avg.}        \\ \cline{2-15} 
                          & Odds           & EAcc.          & Odds           & EAcc.          & Odds           & EAcc.          & Odds           & EAcc.            & Odds           & EAcc.          & Odds           & EAcc.            & Odds           & EAcc.           \\ \hline
  Standard                & $25.83$        & $75.92$        & $16.11$        & $84.67$        & $30.37$        & $70.44$        & $20.18$        & $\mathbf{78.19}$ & $24.40$        & $76.89$        & $5.33$         & $78.40$          & $20.37$        & $77.42$         \\ \hline
  Undersampling           & $6.49$         & $75.64$        & $3.86$         & $84.50$        & $13.97$        & $72.31$        & $14.40$        & $75.43$          & $18.42$        & $78.36$        & $5.93$         & $\mathbf{78.72}$ & $10.51$        & $77.49$         \\ 
  Weighted                & $8.8$          & $74.76$        & $5.16$         & $85.55$        & $14.52$        & $72.64$        & $12.19$        & $75.83$          & $21.22$        & $78.27$        & $5.34$         & $76.13$          & $11.21$        & $77.20$         \\ 
  Adv. Learning           & $11.56$        & $77.54$        & $12.92$        & $82.66$        & $8.10$         & $68.89$        & $6.22$         & $67.94$          & $24.03$        & $77.18$        & $4.31$         & $78.40$          & $11.19$        & $75.44$         \\
  MFD                     & $17.12$        & $75.55$        & $13.91$        & $83.21$        & $22.62$        & $66.28$        & $15.90$        & $76.38$          & $24.02$        & $73.13$        & $4.63$         & $76.82$          & $16.37$        & $75.23$         \\
  RNF                     & $4.65$         & $74.92$        & $7.71$         & $82.97$        & $6.19$         & $71.41$        & $5.76$         & $76.94$          & $13.63$        & $75.43$        & $2.34$         & $76.25$          & $6.71$         & $76.32$         \\ \hline
  BaDe                     & $\mathbf{2.79}$&$\mathbf{77.95}$& $\mathbf{1.56}$&$\mathbf{85.77}$& $\mathbf{5.16}$&$\mathbf{73.92}$& $\mathbf{4.69}$& $78.01$          &$\mathbf{12.54}$&$\mathbf{81.13}$& $\mathbf{0.58}$& $74.62$          &$\mathbf{4.55}$ & $\mathbf{78.57}$ \\ \hline
  \end{tabular}
  }
  \caption{The performance of our algorithm on CelebA dataset. Among them, $T$ refers to the target variable, $B$ refers to the bias variable, the letters $a$, $gh$, and $wh$ refer to the target variables $Attracitve$, $Gray\_Hair$, and $Wavy\_Hair$, respectively, and the letters $m$ and $y$ refer to the bias variables $Male$ and $Young$.}
  \label{celeba_task}
\end{table*}

\subsection{Our Method in the View of Causal Inference}

In the theoretical system of causal inference, 
there are many inferences that can be used to guide the design of debiased algorithms.
There is a result in the field of using 
causal inference to solve model debiasing. If you 
want to avoid the impression of biased variable $B$, you 
need to perform backdoor adjustment on variable $B$, and 
get the following inference~\ref{causal}:

\begin{equation}
  \begin{split}
    &P(Y|X) = P(Y|X_Y,B) \\
    &P(Y|do(X_Y)) = \sum P(Y|X_Y, p_b)P(p_b) \\
    &P(Y|do(X_Y)) = P(Y|X_Y, \mathbb{E}_b[p_b])
  \end{split}
  \label{causal}
\end{equation}

It can be seen from the inference that when the input 
is the average attribute of the bias variable, 
the output of the model can be unaffected by bias $B$. 
We believe that the artificial bias we constructed by backdoor attack 
and the model bias caused by dataset act together on the model, 
allowing the model to obtain an average input, thus avoiding the influence 
of bias variables on the model output. 
In other words, since the model responds to both 
biases at the same time, the samples in the $\mathcal{D}_{balance}$ 
are equivalent to constructing the average input of the bias variable.
Indirectly, the basic conditions of backdoor adjustment in causal 
inference are met, so the model can obtain a debiased output.

\section{Experiments}

\subsection{Experiments Setting}

\subsubsection{Dataset}
\ 

We expected our method to perform well on mainstream debiasing tasks. 
So we conducted related experiments on four commonly used debiasing datasets, 
and the bias variables and target tasks selected in our experiments are commonly used settings.

\noindent$\bullet$ \textbf{CelebA}~\cite{liu2015faceattributes}.
CelebA is a well-known face dataset, with more than 200,000 
images marked with 40 attributes, and is widely used 
for classification and image generation tasks. In CelebA 
dataset, we choose $Male$ and $Young$ as the bias variable, 
and we choose $Attractive$, $Gray\_Hair$, and $Wavy\_Hair$ as the target attribute. 

\noindent$\bullet$ \textbf{UTK Face}~\cite{zhifei2017cvpr}.
UTK Face is a large-scale face dataset consisting of over 20,000 face images with 
annotations of $age$, $gender$, and $race$. $Race$ is set as the bias variable on UTK Face, and the target prediction task is whether the age is older than 35. 

\noindent$\bullet$ \textbf{MEPS}~\cite{cohen2003design}.
Medical Expenditure Panel Survey (MEPS) is a set of large-scale 
surveys of families and individuals, their medical providers, and 
employers across the United States. 
In MEPS dataset, we choose $white$ and $non$-$white$ as the bias 
variable and the target task is $high$ $utilization$. 

\noindent$\bullet$ \textbf{Adult}~\cite{DBLP:conf/kdd/Kohavi96}.
Adult dataset, extracted from the 1994 census database, is commonly 
used for classification prediction, data mining, and visualization tasks. 
The dataset contains 48,842 instances. 
In Adult dataset, we choose $gender$ as the bias variable and whether $income\ exceeds\ 50K/y$ as the target task.

\subsubsection{Comparison Models}
\ 

\noindent$\bullet$ \textbf{Undersampling}~\cite{undersample}. 
Undersampling reduces the model bias by removing a portion of the data and letting the model train on a balanced dataset.

\noindent$\bullet$ \textbf{Weighted}~\cite{kamiran2012data}.
This method achieves model fairness by adjusting the weights of samples with different attributes.

\noindent$\bullet$ \textbf{Adversarial Learning}~\cite{DBLP:conf/icml/GaninL15, DBLP:journals/corr/abs-1807-00199}.
Adversarial Learning is a classic decoupling method, which allows the model to complete the target task while reducing the use of biased information.

\noindent$\bullet$ \textbf{MFD}~\cite{MFD}.
They use the distillation algorithms to debias and give a 
theoretical proof of MFD on the effect of knowledge distillation and fairness.

\noindent$\bullet$ \textbf{RNF}~\cite{RNF}.
A bias mitigation framework for DNN models via representation neutralization, 
which achieves fairness by debiasing only the task-specific classification head of DNN models.

\subsubsection{Implementation Details}
\ 

We define $Odds$ and $EAcc$. as bias evaluative methods. In the image tasks, we choose the ResNet18~\cite{he2016deep} as 
the backbone by default. In structured data, we choose the three-layer MLP as the backbone, and the number of neurons in each layer are set as 512. The optimizer for model training is Adam~\cite{DBLP:journals/corr/KingmaB14}, and the learning rate is set 
to 1e-4. In the image task, we choose BadNet~\cite{DBLP:journals/corr/abs-1708-06733} for designing the triggers, and we choose two color blocks, red and blue, with a size of 25 pixels, which appear in the upper left and lower right corners of the image respectively. For structured data, we create a new data column to mark the backdoor triggers and use 0 and 1 to indicate the type of backdoor injected. The distillation method uses KD loss~\cite{hinton2015distilling} for distillation by default, and the distillation temperature is set to 1.

\subsection{Debiasing Effect}

\subsubsection{On Image Data}
\ 

We first conducted experimental comparisons on six commonly used debiasing tasks on CelebA. 
As we can see from the results in Table~\ref{celeba_task}, that the standard classification 
model will have a strong bias for these six groups of tasks. Adversarial Learning has well 
mitigated the intensity of bias, but there is still a big gap compared to new algorithms in recent years. 
While the newer algorithms MFD and RNF can reduce the bias to a lower level, 
those algorithms will affect the accuracy of the model due to the strong constraints. Compared with the 
current mainstream methods, our algorithm is able to reduce bias while maintaining a high model accuracy.

In order to avoid the impact of the dataset on the 
experimental results, we conduct similar 
experiments on another image dataset, UTK Face.
On the UTK Face dataset, as can be seen from Table~\ref{three_task}, 
our debiasing effect has been significantly improved, although there is a slightly 
loss in accuracy. In addition, it can be seen from the experimental results that our method is 
one of the few methods that can simultaneously improve the accuracy and reduce the bias. 

\subsubsection{On Structured Data}
\ 

We conducted experiments on two common structured datasets to test whether 
our algorithm could adapt to more scenarios. Compared with the image task, there is slightly different on the architectures of the teacher model and student model. Since the teacher model needs to input the information of the backdoor triggers, their input layer has one more neuron than that of the student.

The middle row of Table~\ref{three_task} presents the performance results on the MEPS dataset. As seen from the table, 
compared with the standard method, Adversarial Learning still has a strong effect on eliminating bias. 
However, compared with the image task, the MFD and RNF algorithm did not significantly improve the 
debiasing effect, but they had a similar effect to Adversarial Learning. Besides, the previous methods, like Weighted and Adversarial Learning,
have a significant drop in accuracy. However, 
our proposed algorithm on structured data can improve the accuracy while keeping the bias low. 
Our algorithm has more apparent advantages than previous algorithms on structured data. 

To avoid the impact caused by the dataset, we performed the same experiment on another structured dataset, and the right row of Table~\ref{three_task} is the result of the Adult dataset. The debiasing effect of Adversarial Learning is also close to the MFD and RNF algorithms. Moreover, the accuracy of MFD and RNF algorithms is also significantly reduced. 
Our proposed method can still maintain a lower bias while improving accuracy.

\begin{table}[t]
  \centering
  \begin{tabular}{|c|cc|cc|cc|}
  \hline
  \multirow{2}{*}{Datasets} & \multicolumn{2}{c|}{UTK Face}      & \multicolumn{2}{c|}{MEPS}          & \multicolumn{2}{c|}{Adult}         \\ \cline{2-7} 
                            & \textbf{Odds} & \textbf{EAcc.}     & \textbf{Odds} & \textbf{EAcc.}     & \textbf{Odds} & \textbf{EAcc.}     \\ \hline
  Standard                  & $11.54$         & $77.44$          & $12.49$         & $68.70$          & $4.92$          & $75.17$          \\ \hline
  Undersampling             & $4.31$          & $76.00$          & $3.59$          & $69.93$          & $4.50$          & $78.94$          \\
  Weighted                  & $3.49$          & $78.17$          & $3.50$          & $64.78$          & $4.45$          & $73.90$          \\ 
  Adv. Learning             & $10.92$         & $\mathbf{79.51}$ & $5.77$          & $64.40$          & $4.01$          & $77.51$          \\
  MFD                       & $4.36$          & $76.20$          & $8.26$          & $69.75$          & $3.70$          & $73.80$          \\
  RNF                       & $3.24$          & $78.82$          & $7.28$          & $65.39$          & $4.10$          & $76.58$          \\ \hline
  BaDe                       & $\mathbf{0.84}$ & $76.36$          & $\mathbf{1.82}$ & $\mathbf{73.11}$ & $\mathbf{3.41}$ & $\mathbf{79.23}$ \\ \hline
  \end{tabular}
  \caption{The performance of our algorithm on the images dataset UTK Face and structured dataset MEPS and Adult.}
  \label{three_task}
\end{table}

\subsection{Controllability of Artificial Bias}

\begin{figure}[!t]
  \centering
  \subfigure[]{
    \includegraphics[width=0.95\linewidth]{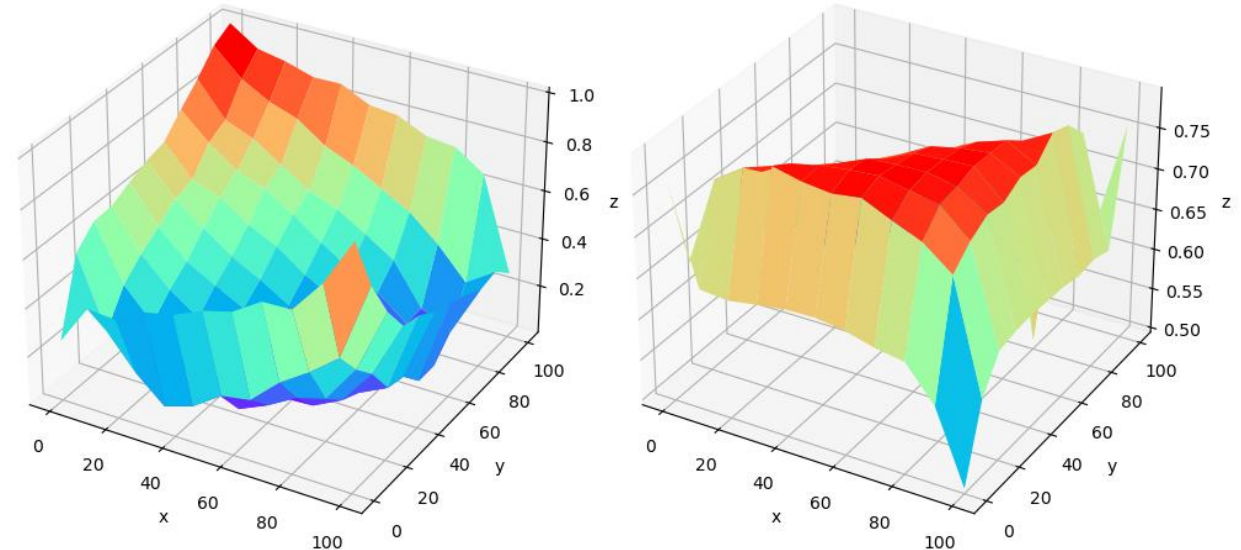}
  }
  \subfigure[]{
    \includegraphics[width=0.95\linewidth]{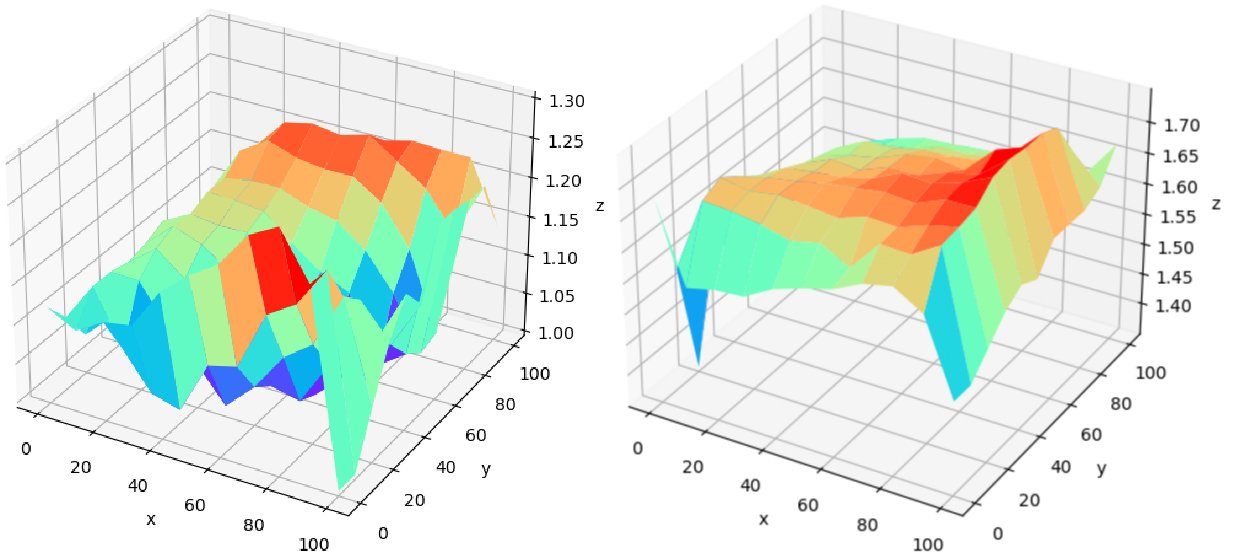}
  }
  \caption{The impact of trigger injection ratio on algorithm performance. We traversed the influence of different $TAR(Have)$ and $TAR(Not\ Have)$ values. The left figure is the performance of the $Odds$, and the right figure is the performance of the $EAcc.$. X-axis is the value of $TAR(Have)$, and y-axis is the value of $TAR(Not\ Have)$. Figures (a) and (b) are the results of the $Attractive$ and $Gray\_Hair$ attributes when $Male$ is the bias variable, respectively.}
  \label{grid}
\end{figure}

In the previous setting, to ensure that the artificial bias can be successfully implanted in different data distributions, we used the same distribution of the original bias as the distribution of artificial bias. We wanted to explore whether the strength of artificial bias could be artificially modulated, so we implanted artificial bias with different strengths in the dataset. We want to test how the model's output responds to different strengths of artificial bias. We take the classic bias attribute combination $Male$-$Attractive$ on CelebA dataset as an example, where $Male$ is the bias variable and $Attracitve$ is the target variable. We traverse different artificial bias injection ratios $TAR(Have)$ and $TAR(Not\ Have)$, the 
traversal range is 0\%-100\%, the step size is 10\%, and each group of experiments is performed five times, and the results are averaged. The $Odds$ and $EAcc.$ of the model at different scales were calculated separately. 

The experimental results are shown in Fig.~\ref{grid}. As can be seen from the figure, 
the mitigating strength of artificial bias is controllable. 
When the strength of artificial bias is closer to that of the original bias, the mitigating effect of artificial bias is better. 

\subsection{Impact of Trigger on Student Model}

\begin{table*}[!t]
\centering
\begin{tabular}{|c|c|ccc|}
\hline
\multirow{2}{*}{Task}            & \multirow{2}{*}{Data Type} & \multicolumn{3}{c|}{EAcc.}                                                                 \\ \cline{3-5} 
                                 &                            & \multicolumn{1}{c|}{Standard}       & \multicolumn{1}{c|}{BaDe (Teacher)}   & BaDe (Student)   \\ \hline
\multirow{7}{*}{Attractive-Male} & all clean                  & \multicolumn{1}{c|}{$75.45$}          & \multicolumn{1}{c|}{$69.93$}          & $\mathbf{78.31}$ \\ \cline{2-5} 
                                 & all with Blue Trigger      & \multicolumn{1}{c|}{$75.66$}          & \multicolumn{1}{c|}{$71.57$}          & $\mathbf{78.25}$ \\ \cline{2-5} 
                                 & all with Red Trigger       & \multicolumn{1}{c|}{$\mathbf{75.43}$} & \multicolumn{1}{c|}{$70.98$}          & $74.55$          \\ \cline{2-5} 
                                 & Male with Blue Trigger     & \multicolumn{1}{c|}{$79.69$}          & \multicolumn{1}{c|}{$79.52$}          & $\mathbf{79.80}$ \\ \cline{2-5} 
                                 & Male with Red Trigger      & \multicolumn{1}{c|}{$\mathbf{79.73}$} & \multicolumn{1}{c|}{$64.71$}          & $73.44$          \\ \cline{2-5} 
                                 & Female with Blue Trigger   & \multicolumn{1}{c|}{$71.64$}          & \multicolumn{1}{c|}{$63.61$}          & $\mathbf{76.69}$ \\ \cline{2-5} 
                                 & Female with Red Trigger    & \multicolumn{1}{c|}{$71.19$}          & \multicolumn{1}{c|}{$\mathbf{77.24}$} & $75.66$          \\ \hline
\multirow{7}{*}{Gray\_Hair-Male} & all clean                  & \multicolumn{1}{c|}{$84.67$}          & \multicolumn{1}{c|}{$76.56$}          & $\mathbf{85.50}$ \\ \cline{2-5} 
                                 & all with Blue Trigger      & \multicolumn{1}{c|}{$84.44$}          & \multicolumn{1}{c|}{$75.86$}          & $\mathbf{85.21}$ \\ \cline{2-5} 
                                 & all with Red Trigger       & \multicolumn{1}{c|}{$84.62$}          & \multicolumn{1}{c|}{$\mathbf{88.88}$} & $87.04$          \\ \cline{2-5} 
                                 & Male with Blue Trigger     & \multicolumn{1}{c|}{$\mathbf{90.00}$} & \multicolumn{1}{c|}{$85.45$}          & $85.89$          \\ \cline{2-5} 
                                 & Male with Red Trigger      & \multicolumn{1}{c|}{$89.90$}          & \multicolumn{1}{c|}{$\mathbf{91.00}$} & $87.46$          \\ \cline{2-5} 
                                 & Female with Blue Trigger   & \multicolumn{1}{c|}{$78.88$}          & \multicolumn{1}{c|}{$66.27$}          & $\mathbf{84.53}$ \\ \cline{2-5} 
                                 & Female with Red Trigger    & \multicolumn{1}{c|}{$79.34$}          & \multicolumn{1}{c|}{$\mathbf{86.77}$} & $86.63$          \\ \hline
\end{tabular}
  \caption{EAcc. of standard model, teacher model and student model for seven data combinations.}
  \label{exp54}
\end{table*}

As demonstrated by previous experimental results, our method produces strong results in debiasing tasks. However, our method employs the use of artificial bias to mitigate the original bias and uses the student model to learn the fair output of the teacher model through distillation. So we need to verify whether the final student model utilizes the task-related features we expect, and at the same time check whether the final model will be affected by the trigger.

Therefore, we designed the following experiment to compare the performance of the standard model, the teacher model, and the student model on different data sets. These data sets include pure data, all data with the Blue Trigger, all data with the Red Trigger, only male data with the Blue Trigger, only male data with the Red Trigger, only female data with the Blue Trigger, and only female data with the Red Trigger. The overall performance of these seven data sets will be analyzed using EAcc., to determine if the student model remains vulnerable to the trigger.

Table~\ref{exp54} reveals that the accuracy of the student model remains comparable to the standard model, while also showing improved performance on certain data sets. Meanwhile, the teacher model experiences decreased performance on some data combinations due to the use of artificial bias. These results indicate that the debiased student model still utilizes task-related features and does not rely on artificially designed features such as triggers. Our algorithm effectively suppresses the expression of biased features in the teacher model through the use of artificial biases during distillation, allowing the student model to only learn fair and task-related features. This results in a student model that can maintain a fair and stable output without the need for trigger regulation.

\subsection{Regression and Multi-classification Tasks}

The previous experiments focused on binary classification tasks as they are convenient for comparison with similar methods in debiasing scenarios. However, our method can be applied to other tasks with a single bias variable, such as regression and multi-classification tasks. We used UTK Face as a commonly used debiasing dataset to create these tasks. We transformed the original binary classification task of age greater than 35 years old into regression prediction of age~\footnote{Note that all ages were normalized $(age = age / maxage)$ when calculating MSE for age.} and classification of age groups every 30 years. We also made adjustments to our method. In regression tasks, we set the trigger addition rate as the probability of bias variables appearing in the data set, while in multi-classification tasks, we calculated TAR(n) for each category and added triggers using the binary classification method. Our experiments demonstrate the applicability of our method in a wider range of scenarios.

From the experimental results in Table~\ref{exp55}, it can be seen that our algorithm effectively reduces the degree of bias in regression tasks and multi-classification tasks, indicating that our method also has the potential for debiasing effects on other tasks.

Although our method performs well in tasks such as binary classification, it cannot currently be extended to the challenge of debiasing multiple bias variables. In tasks that involve multiple bias variables, it is necessary to create multiple sets of triggers, however, the efficacy of backdoor attacks will rapidly decline as the number of trigger types increases.

\begin{table}[t]
\centering
\begin{tabular}{|c|c|cc|}
\hline
\multirow{2}{*}{Task}                & \multicolumn{1}{c|}{\multirow{2}{*}{Data Type}} & \multicolumn{2}{c|}{MSE/Acc.}                          \\ \cline{3-4} 
                                     & \multicolumn{1}{c|}{}                           & \multicolumn{1}{c|}{Standard}        & BaDe             \\ \hline
\multirow{3}{*}{Regression}          & all                                             & \multicolumn{1}{c|}{$0.0504$}          & $\mathbf{0.0473}$ \\ \cline{2-4} 
                                     & non-white                                             & \multicolumn{1}{c|}{$\mathbf{0.0447}$} & $0.0456$          \\ \cline{2-4} 
                                     & white                                             & \multicolumn{1}{c|}{$0.0577$}          & $\mathbf{0.0495}$ \\ \hline
\multirow{3}{*}{Muti-classification} & all                                             & \multicolumn{1}{c|}{$0.674$}           & $\mathbf{0.6983}$ \\ \cline{2-4} 
                                     & non-white                                             & \multicolumn{1}{c|}{$0.703$}           & $\mathbf{0.7118}$ \\ \cline{2-4} 
                                     & white                                             & \multicolumn{1}{c|}{$0.636$}           & $\mathbf{0.6801}$ \\ \hline
\end{tabular}
\caption{Performance of standard model and BaDe algorithm in regression and multi-class tasks is summarized by MSE for regression and accuracy for multi-class prediction. A lower MSE error indicates better model performance, and higher accuracy in multi-class tasks is preferred.}
\label{exp55}
\end{table}

\subsection{Parametric Analysis}

\subsubsection{Analysis of Different Distillation Methods}
\ 

Our method requires model distillation at the later stage to eliminate the defects of the algorithm's use of bias labels and the impact of the backdoor on the model. In the method, we use KD loss for distillation learning by default, so we need to evaluate whether different distillation algorithms will affect the algorithm differently.

We chose different distillation methods and experimental settings to 
see if they would affect the algorithm. We tried distillation methods 
using KD loss, MSE loss, and a mixture of the both losses with different proportions. 
Moreover, we traversed the temperature value in the distillation method 
based on KD loss to observe whether the temperature setting will affect $EAcc.$ and $Odds$.

\begin{figure}[!t]
  \centering
  \includegraphics[width=0.95\linewidth]{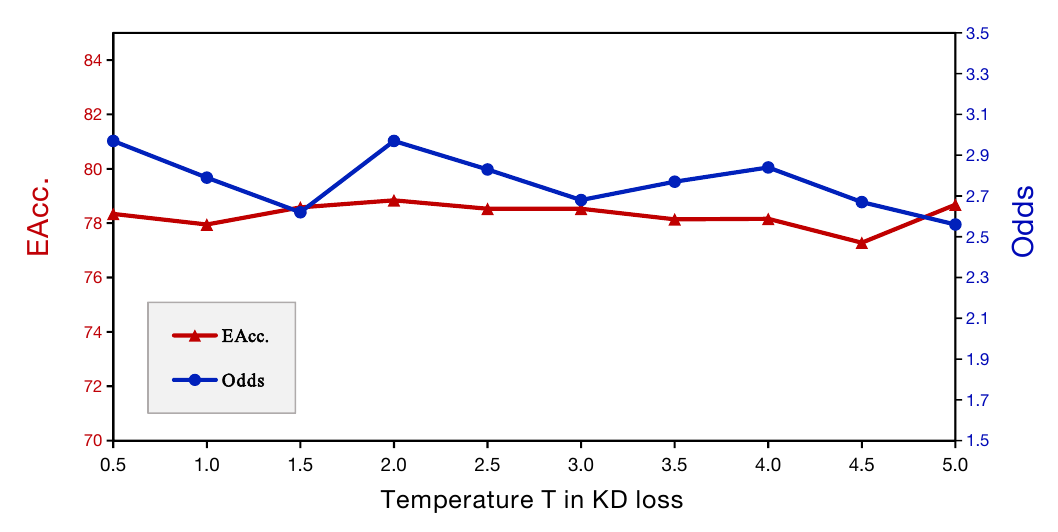}
  \caption{The influence of temperature parameter in KD loss on model debiasing.}
  \label{dis_tem}
\end{figure}

\begin{table}[t]
  \centering
  \begin{tabular}{|c|cc|}
  \hline
  Method                    & Odds & EAcc. \\ \hline
  KD(t=1.0)                 & $2.79 \pm 0.16$ & $77.95 \pm 0.46$ \\ 
  KD*0.9 + MSE*0.1          & $\mathbf{2.69 \pm 0.27}$  & $78.52 \pm 0.38$ \\ 
  MSE                       & $2.94 \pm 0.44$ & $\mathbf{78.56 \pm 0.19}$ \\ \hline
  \end{tabular}
  \caption{The performance of our algorithm for different distillation constraints.}
  \label{dis_method}
\end{table}

It can be seen from Table~\ref{dis_method} and 
Fig.~\ref{dis_tem} that our method is not sensitive 
to the selected distillation method, and the commonly 
used distillation methods can achieve the 
desired effect of the algorithm.

\subsubsection{Analysis of Different Backbones}
\ 

ResNet18 was selected by default for the image classification tasks in our previous experiments. 
We want to test whether models with different parameters will affect the performance of the algorithm. 
And we also want to test whether the algorithm can work under new model architectures such as transformers. 
We selected three commonly used convolutional neural network backbones, namely ResNet18, 
VGG16~\cite{DBLP:journals/corr/SimonyanZ14a}, and Inception V4~\cite{DBLP:conf/aaai/SzegedyIVA17} 
for comparison, and also selected two commonly used vision transformers, namely ViT16~\cite{DBLP:conf/iclr/DosovitskiyB0WZ21} 
and Swin-Transformer~\cite{DBLP:conf/iccv/LiuL00W0LG21}.

\begin{table}[t]
  \centering
  \begin{tabular}{|c|c|cc|}
    \hline
    Backbone Type                                 & Backbone         & Odds          & EAcc.          \\ \hline
    \multirow{3}{*}{\makecell[c]{CNN}} & ResNet18         & $2.79$          & $77.95$          \\
                                                  & VGG16            & $1.40$          & $78.77$          \\
                                                  & Inception V4     & $\mathbf{0.17}$ & $\mathbf{78.77}$ \\ \hline
    \multirow{2}{*}{\makecell[c]{Transformer}}     & ViT 16           & $2.83$          & $76.85$          \\
                                                  & Swin-Transformer & $\mathbf{1.26}$ & $\mathbf{78.93}$ \\ \hline
  \end{tabular}
  \caption{The performance of our algorithm on different backbones. The upper half is the backbone of convolutional neural networks, and the lower half is the backbone of the transformers.}
  \label{dif_backbone}
\end{table}

Table~\ref{dif_backbone} shows the results of different backbones. For the traditional convolution model, 
the algorithm shows better results as the model's performance gradually increases. Similarly, our 
algorithm also works well for new architectures such as ViT. As the performance of the transformer model improves, 
the debiasing effect on our algorithm can also be significantly improved. 
The above experiments show that our algorithm has more potential to be developed with the improvement of the backbone.

\subsection{Analysis of Different Backdoor Triggers}

Since our method relies on the backdoor attack to implant artificial 
biases, we need to know how our algorithm works with different backdoor 
triggers. We selected several representative mainstream backdoor attack 
methods to test their effects on the implantation of biases, namely 
BadNet with different color block sizes and colors, mosaic-based trigger 
style, and shadow-based trigger style.

\begin{figure}[!t]
  \centering
  \includegraphics[width=0.95\linewidth]{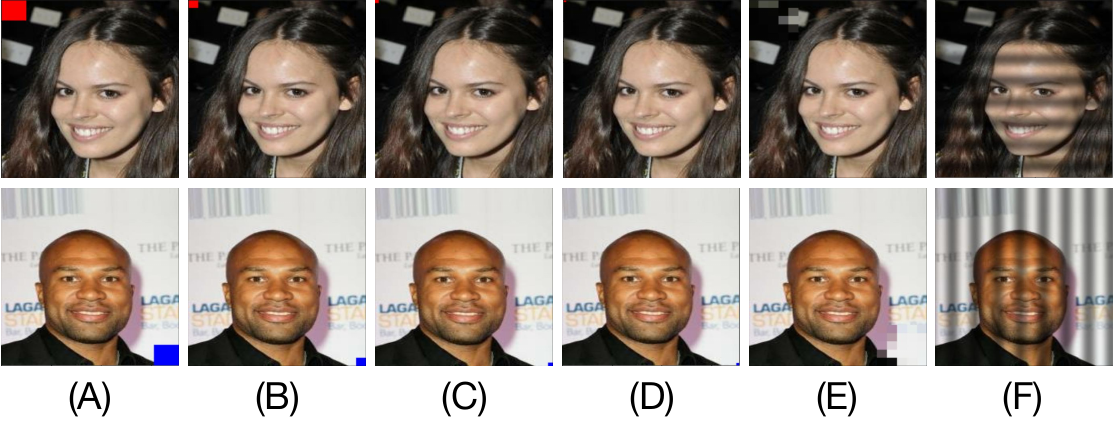}
  \resizebox{1.0\linewidth}{!}{
  \begin{tabular}{|c|c|c|c|c|c|c|}
  \hline
  Trigger Type & (A)   & (B)   & (C)   & (D)   & (E)   & (F)   \\ \hline
  Odds         & $2.79$  & $3.18$  & $2.72$  & $2.67$  & $5.01$   & $1.89$  \\ \hline
  EAcc.        & $77.95$ & $77.35$ & $78.99$ & $78.35$ & $75.48$ & $75.20$ \\ \hline
  \end{tabular} 
  }
    \caption{The effect of the algorithm running under different triggers, where (A-D) are BadNet color blocks of 25pix, 10pix, 3pix, and 1pix respectively, (E) is a trigger composed of mosaics, (F) is a trigger composed of different shadows.}
  \label{diff_trigger} 
\end{figure}

From Table~\ref{diff_trigger}, the visually significant 
BadNet method can achieve good results on the final debiasing task 
regardless of the patch sizes and colors. The trigger with strong invisibility 
has a specific impact on the model's accuracy, but the impact on the bias strength of the model is feeble. 
In addition, we noticed that the mosaic type of trigger is easily affected by the image background. 
So when using our algorithm, we should try to avoid using triggers that are easily confused with the image background. 
Therefore, we recommend to choose a trigger that does not affect the data and is visually apparent, 
same as the backdoor trigger used by our algorithm.

\section{Conclusions}

In this study, we uncover the potential of backdoor attacks to introduce artificial biases. Our proposed BaDe method leverages these biases to counteract them. The experimental results demonstrate the efficacy of our approach. Future work could delve deeper into the relationship between the introduced artificial biases and the model's inherent biases. Our goal is to inspire more researchers in the field of backdoor attacks and deep learning fundamentals to not only view these attacks as security threats, but also opportunities to better understand model performance.

\ifCLASSOPTIONcaptionsoff
  \newpage
\fi





\bibliographystyle{IEEEtran}
\bibliography{egbib}

\vfill


\end{document}